\documentclass[11pt]{article}
\usepackage[utf8]{inputenc}
\usepackage{amsmath, amssymb, amsthm}
\theoremstyle{plain}
\newtheorem{theorem}{Theorem}[section] % or omit [section] if you don't want per-section numbers

\usepackage{hyperref}
\theoremstyle{definition}
\newtheorem{definition}[theorem]{Definition}

\theoremstyle{remark}

\usepackage{pgfplots}
\pgfplotsset{compat=1.18}
\usetikzlibrary{arrows.meta,positioning}

\usepackage{times}
\usepackage{fullpage}

\newcommand{\ignore}[1]{}

\def\bold0{\mathbf{0}}

\def\epsilon{\varepsilon}

\newtheorem*{theorem*}{Theorem}

\newtheorem*{conjecture*}{Conjecture}

\newtheorem*{definition*}{Definition}

\newtheorem*{rep@theorem}{\rep@title}
\newcommand{\newreptheorem}[2]{%
\newenvironment{rep#1}[1]{%
 \def\rep@title{#2 \ref{##1}}%
 \begin{rep@theorem}}%
 {\end{rep@theorem}}}

\newreptheorem{theorem}{Theorem}
\newreptheorem{lemma}{Lemma}
\newreptheorem{proposition}{Proposition}
\newreptheorem{claim}{Claim}
\newreptheorem{corollary}{Corollary}
\newreptheorem{mainlemma}{Main Lemma}

\theoremstyle{remark}

\title{Research Program: \\ Theory of Learning in Dynamical Systems}
\author{Elad Hazan  \and Shai Shalev Shwartz  \and  Nathan Srebro}
\date{}

\begin{document}
\maketitle

\begin{abstract}
Modern learning systems increasingly interact with data that evolve over time and depend on hidden internal state. We ask a basic question: when is such a dynamical system  learnable from observations alone?

This paper proposes a research program for understanding learnability in dynamical systems through the lens of next-token prediction.  We argue that learnability in dynamical systems should be studied as a finite-sample question, and be based on the properties of the underlying dynamics rather than the statistical properties of the resulting sequence. 

To this end, we give a formulation of  learnability for stochastic processes induced by dynamical systems, focusing on guarantees that hold uniformly at every time step after a finite burn-in period. This leads to a notion of dynamic learnability which captures how the structure of a system—such as stability, mixing, observability, and spectral properties—governs the number of observations required before reliable prediction becomes possible. 

We illustrate the framework in the case of linear dynamical systems, showing that accurate prediction can be achieved after finite observation without system identification, by leveraging improper methods based on spectral filtering. We survey the relationship between learning in dynamical systems and classical PAC, online, and universal prediction theories, and suggest directions for studying nonlinear and controlled systems.

\end{abstract}

\section{Introduction}
\label{sec:introduction}

Classical statistical learning theory (e.g., PAC learning) \cite{Valiant1984, Kearns1994, Blumer1989, shalev2014understanding}
provides a foundational framework for reasoning about sample complexity in static settings.
A hypothesis class $\mathcal{H}$ is said to be learnable if, given i.i.d.\ samples from a fixed distribution,
one can find a hypothesis with low generalization error.
This view was successful for tasks such as classification and regression, but it assumes that data are generated
i.i.d.\ from a static distribution and that the learner competes with a fixed hypothesis.

Modern artificial intelligence, however, is inherently sequential and dynamical.
Brains, language models, and control systems all operate in environments with state and memory.
Data are emitted as evolving sequences, not as independent samples.
This raises a fundamental question that is not naturally addressed by classical learning theory:
\emph{when does finite observation suffice to make reliable predictions in dynamical systems with hidden state?}

The purpose of this note is to propose a research agenda of sequential prediction. The prediction methodology should have two main properties. First, it should consider finite samples rather than asymptotic bounds. Secondly, it should emphasize the complexity of a class of sequences based on the dynamics that generated them, rather than statistical properties of the observation sequence.  We describe such a framework henceforth.

\vspace{0.5em}
\noindent
\textbf{Sequential prediction and its limitations.}
The problem of predicting sequences has been studied extensively in the theory of {universal prediction}
and online learning.
In particular, the work of Merhav and Feder \cite{merhav2002universal} develops a comprehensive theory of
universal prediction for stochastic and individual sequences, providing asymptotic and time-averaged guarantees
relative to an optimal predictor or Bayes envelope.
Closely related, Vovk and others \cite{Vovk1990, Vovk1995} formulate prediction as a game and study algorithms
that compete with a rich class of predictors under adversarial or arbitrary sequences, often with strong
pathwise regret guarantees. The notion of regret was also applied to time series prediction in the online learning literature \cite{anava2013online,kuznetsov2016time}.

While these frameworks are powerful, they do not address a central aspect of learning in dynamical systems.
They treat the environment primarily as an abstract sequence or source, rather than as a system with latent state,
state evolution, and structural properties such as stability, mixing, or observability.
As a result, they focus on asymptotic optimality or average regret, but do not ask whether
finite observation resolves hidden-state uncertainty well enough to enable reliable long-term prediction,
nor how this depends on the complexity of a class of dynamical systems.

\vspace{0.5em}
\noindent
\textbf{Learning in dynamical systems as a finite-sample problem.}
This position paper argues that learnability in dynamical systems should be studied as a
\emph{finite-sample question at the level of a class of systems}:
how many observations are required before accurate prediction becomes possible,
and how does this depend on the structure of the dynamics?
To address this question, we propose a dynamical analog of PAC and online learnability,
tailored to prediction with memory and hidden state.

\paragraph{What we mean by dynamic learnability.}
A class of systems is {dynamically learnable} if there exists a single predictor
and a sample complexity $T(\varepsilon)$ such that, for every admissible initial state $x_{0}$,
the per-step {prediction risk}
---defined as the expected loss $\mathbb{E}[\ell(\hat{y}_{t+1}, y_{t+1})]$
under a suitable task-specific loss function $\ell : \mathcal{Y} \times \mathcal{Y} \to \mathbb{R}_{+}$---
is at most $\varepsilon$ for all times $t \ge T(\varepsilon)$.
Formally, the notion of risk will be defined in Section~\ref{sec:learnability-conditions}.

This definition emphasizes three features that are intrinsic to learning in dynamical systems:
(i) a \emph{finite-sample} notion of learnability through the burn-in time $T(\varepsilon)$,
(ii) explicit treatment of \emph{dynamical systems with hidden state},
and (iii) guarantees that hold \emph{uniformly over time} after the burn-in, rather than only on average. While uniform-in-time guarantees may be stronger than necessary in some applications, they provide a clean notion of having resolved hidden-state uncertainty. 

\paragraph{Departures from classical learning theory.}
The fundamental distinction of the proposed agenda is that learnability is governed by the {structure of the generating dynamics}---such as stability, mixing, observability, and spectral properties---rather than the statistical properties of the observed sequence alone.

This perspective differs from classical statistical (PAC) learning, which assumes i.i.d.\ samples from a fixed distribution. While one could technically reduce sequential prediction to a PAC-style setting by ``unrolling'' the dynamical system into a distribution over length-$T$ trajectories, such a reduction is unnatural. It obscures temporal structure, ignores the role of latent state, and treats prediction as if it were independent sampling rather than evolution over time.

Similarly, this perspective differs from online learning \cite{CesaBianchi2006, Hazan2022, orabona2019modern, ShalevShwartz2012}, which typically treats the observation sequence as adversarial or arbitrary. While online learning minimizes regret (average performance), it does not address whether the learner resolves the hidden-state uncertainty. In contrast, dynamic learnability exploits the constraints of the underlying system to ensure that, after a finite burn-in time $T(\varepsilon)$, the risk remains small for every future time step.

\paragraph{Bridging learning theory and control.}
This proposal aims to bridge the gap between distinct traditions in modeling dynamics.
Classically, system identification \cite{ljung1999system} has focused on the \emph{asymptotic} consistency of estimating model parameters.
More recently, these ideas have evolved into "operator learning" for PDEs and infinite-dimensional systems \cite{kovachki2021neural, lu2021learning}, which learns mappings between function spaces rather than finite-dimensional states.
Parallel to this, the theory of Koopman operators \cite{mezic2005spectral, brunton2021modern} provides a powerful toolkit for analyzing nonlinear dynamics through linear spectral decompositions.
However, while these lines of work provide the \emph{representations} and approximation theories, they typically do not address the statistical difficulty of the learning task.
The contribution of the present work is to apply the finite-sample methodology of learning theory to these structures, asking not just whether a system can be represented, but what is the \emph{sample complexity} required to resolve its hidden state for reliable prediction.

\paragraph{Why is next-token prediction a good metric?}
Our formulation is motivated by the empirical success of next-token prediction,
most notably in large language models.
Rather than explaining this phenomenon, we take next-token prediction as a canonical task
and ask when it is \emph{possible} in systems with hidden state,
and how its feasibility depends on the structure and complexity of the underlying dynamics.

\section{Improper Dynamic Learnability without Identification} \label{sec:case_study}

In this section we describe a motivational example for our definitions: a simple variant of learning linear dynamical systems. This case study brings out the main advantages of the definitions: allowing improper learning, and not requiring system identification, for learning in dynamical systems.

Consider the well-studied setting of \emph{linear dynamical systems (LDS)}.
A linear dynamical system (with no control inputs) evolves as
\begin{align*}
    x_{t+1} &= A x_t + w_t, \\
    y_t &= C x_t + v_t,
\end{align*}
where $x_t \in \mathbb{R}^d$ is the hidden state, $y_t \in \mathbb{R}^p$ is the observation, and
$(w_t), (v_t)$ are mean-zero noise processes. We also consider the case that $A$ is symmetric, for ease of exposition, although extensions are possible to general LDS and even certain nonlinear dynamics \cite{marsden2025universal,Dogariu2025}. Classical system identification asks whether one can
estimate $(A,C)$, while in our setting the question is whether the sequence $(y_t)$ is
\emph{dynamically learnable}.

\subsection{Spectral Filtering for Symmetric Linear Dynamical Systems}
\label{subsec:symmetric-spectral-filtering}

We now describe in detail the spectral filtering approach
for learning symmetric linear dynamical systems from~\cite{hazan2017learning}.
This provides a concrete and elegant example of
\emph{improper dynamic learnability}—accurate prediction
without system identification. % We note that while the presentation is given for symmetric LDS for simplicity, recent work has extended the approach for general LDS and even certain nonlinear dynamical systems \cite{marsden2025universal,Dogariu2025}.

\paragraph{Symmetric linear dynamical systems.}
Consider the LDS
\begin{equation}
x_{t+1} = A x_t + w_t, \qquad
y_t = C x_t + v_t,
\label{eq:symmetric-lds}
\end{equation}
where $A \in \mathbb{R}^{d\times d}$ is symmetric with $\|A\|_2 \le 1$,
$C \in \mathbb{R}^{1\times d}$, and $(w_t),(v_t)$ are mean-zero
sub-Gaussian noise processes independent across time.
By symmetry, $A$ admits an orthogonal eigendecomposition
$A = U \Lambda U^\top$ with $\Lambda = \operatorname{diag}(\lambda_1,\ldots,\lambda_d)$,
$\lambda_i \in [-1,1]$.
Unrolling~\eqref{eq:symmetric-lds} gives the expected output
\[
\mathbb E[y_{t+1} \mid y_{1:t}]
= \sum_{k=1}^\infty \beta_k y_{t+1-k}, \qquad
\beta_k = C A^{k-1} C^\top
= \sum_{i=1}^d (C u_i)^2 \lambda_i^{k-1}.
\]
Thus, the Bayes--optimal predictor is an infinite convolution of past outputs
with coefficients $(\beta_k)$ that are smooth mixtures of exponentials
$\lambda^k$ for $\lambda \in [-1,1]$.

\paragraph{Universal Spectral Basis.}
Spectral filtering replaces the unknown kernel $(\beta_k)$ by a small
bank of fixed, data-independent filters that approximate all exponentials
$\lambda^k$ uniformly over $\lambda \in [-1,1]$.
Let $\mathsf{H}_T \in \mathbb{R}^{T\times T}$ be the Hilbert matrix
$(\mathsf{H}_T)_{ij} = 1/(i+j-1)$, and let
$\phi_1, \ldots, \phi_m$ be its top $m$ orthonormal eigenvectors,
ordered by eigenvalue.
Each $\phi_j$ defines a convolutional feature
\[
z_t^{(j)} = \sum_{k=1}^{T} \phi_{j,k}\, y_{t+1-k},
\qquad
z_t = (z_t^{(1)},\ldots,z_t^{(m)}),
\]
and the predictor is linear:
$\hat y_{t+1} = w^\top z_t$.
The readout weights $w$ are learned by least squares on the trajectory
$\{(z_t,y_{t+1})\}$.

\begin{figure}[t]
    \centering
    \includegraphics[width=0.5\linewidth]{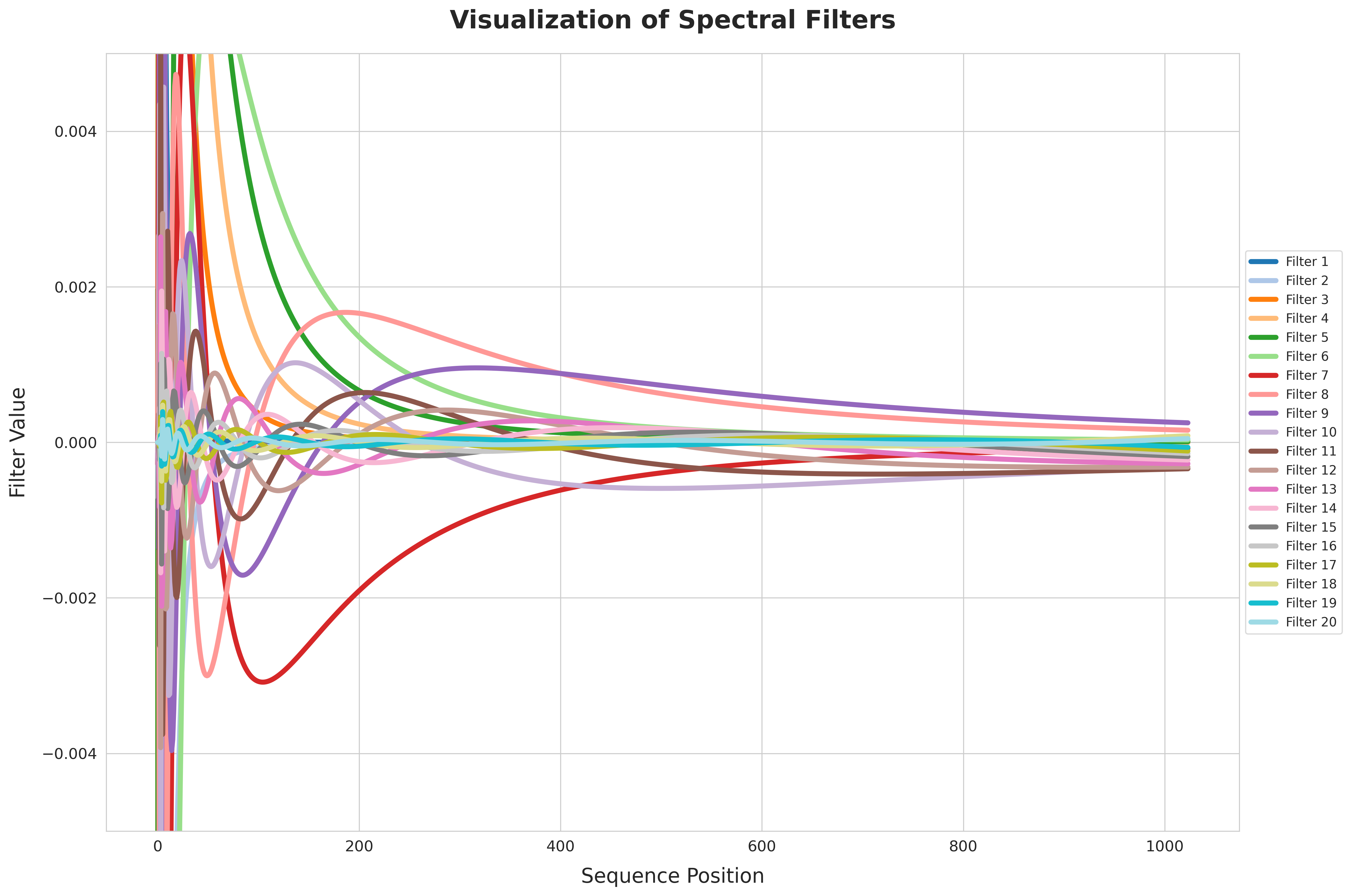}
    \caption{
        Example spectral filters $\phi_1,\phi_3,\phi_5,\phi_{10},\phi_{20}$
        obtained as the top eigenvectors of a Hilbert or Hankel matrix.
            }
    \label{fig:filters}
\end{figure}

\begin{theorem}[Dynamic Learnability of Symmetric LDS]
\label{thm:symmetric-sf}
Let the system~\eqref{eq:symmetric-lds} satisfy $\|A\|_2 \le 1$ and be driven by independent sub-Gaussian noise.
Let $\phi_1,\ldots,\phi_m$ be the top $m$ eigenvectors of
the Hilbert matrix $\mathsf{H}_T$.
There exists a choice of filter count $m = \Theta(\log T \log(1/\varepsilon))$ such that the spectral filtering algorithm $\mathcal{A}$ satisfies the definition of Dynamic Learnability (Def \ref{def:ds-learnability}) with complexity:
\[
T(\varepsilon) = \tilde{O}\left( \frac{1}{\varepsilon} \right).
\]
Specifically, for all $t > T(\varepsilon)$, the expected excess risk is bounded by:
\[
\mathbb E\!\left[
  \| \hat{y}_{t+1} - y_{t+1} \|^2 - \| \pi^\star_{P}(y_{1:t}) - y_{t+1} \|^2 
\right]
\le \varepsilon.
\]
\end{theorem}

\begin{proof}[Proof Sketch of Theorem \ref{thm:symmetric-sf}]
We decompose the instantaneous risk at time $t$ into approximation (bias) and estimation (variance) errors. 
Let $w^* = \arg\min_w \mathbb{E}[(w^\top z_t - y_{t+1})^2]$ denote the optimal linear predictor in the spectral feature space.
By the standard bias-variance decomposition for squared loss, the expected excess risk at time $t$ can be written as the sum of two independent terms:
\[
\underbrace{\mathbb{E}[((\hat{w}_t - w^*)^\top z_t)^2]}_{\text{Estimation Error (Variance)}}
+
\underbrace{\mathbb{E}[(w^{*\top} z_t - \pi^\star_P(Y_{1:t}))^2]}_{\text{Approximation Error (Bias)}}.
\]
We bound each term separately:
\begin{enumerate}
    \item \textbf{Approximation Error (Bias):}
    The Bayes-optimal predictor $\pi^\star_P$ is realized by an infinite convolution kernel $\beta = (\beta_k)$, which is a convex mixture of exponentials $\lambda^k$. The ability of the filters to approximate this is captured by the eigenvalues of the Hilbert matrix. Specifically, if we denote by $\mu_1,\dots,\mu_T$ the eigenvalues of the Hilbert matrix $\mathsf{H}_T$, and consider an impulse response vector $v_\lambda = (1, \lambda, \dots, \lambda^{T-1})^\top$, then:
    \begin{align*}
    \mathbb{E}_{\lambda \in [0,1]} [(v_\lambda^\top \phi_i)^2] & = \phi_i^\top  \mathbb{E}_{\lambda \in [0,1]} [ v_\lambda v_\lambda^\top  ] \phi_i \\
    & = \phi_i^\top  \int_{0}^{1} \begin{pmatrix}
    1 & \lambda & \cdots & \lambda^{T-1} \\
    \lambda & \lambda^2 & \cdots & \lambda^{T} \\
    \vdots & \vdots & \ddots & \vdots \\
    \lambda^{T-1} & \lambda^{T} & \cdots & \lambda^{2T-2}
    \end{pmatrix} d\lambda \, \phi_i \\
    & = \phi_i^\top \mathsf{H}_T \phi_i = \mu_i.
    \end{align*}
    Using the result from \cite{beckermann2017singular} that $\mu_i \approx \exp(-\frac{\pi^2 i}{2 \log T})$, the energy of the system dynamics falling outside the top $m$ filters is exponentially small. For $m = \Theta(\log T \log(1/\varepsilon))$, the bias is bounded by $\varepsilon/2$.

\item \textbf{Estimation Error (Variance):}
    The system is driven by stochastic noise $(w_t, v_t)$, which acts as a source of \emph{persistent excitation} for the spectral features $z_t$.
    This stochasticity ensures that the covariance matrix $\Sigma_t = \sum_{\tau=1}^t z_\tau z_\tau^\top$ is well-conditioned, satisfying $\lambda_{\min}(\Sigma_t) \ge \Omega(t)$ with high probability.
    Standard results for Least Squares in stochastic settings (e.g., \cite{lai1982least}) then bound the parameter error in the Mahalanobis norm:
    \[ \|\hat{w}_t - w^*\|_{\Sigma_t}^2 = (\hat{w}_t - w^*)^\top \Sigma_t (\hat{w}_t - w^*) \le O(d \sigma^2 \log t). \]
    Since the expected instantaneous excess risk at time $t$ is proportional to $\frac{1}{t} \|\hat{w}_t - w^*\|_{\Sigma_t}^2$, the variance term decays as:
    \[ \mathbb{E}[\text{Variance}_t] \le O\left(\frac{m \log t}{t}\right). \]

\end{enumerate}
To achieve total risk $\le \varepsilon$, we require the variance term $O(m/t) \le \varepsilon/2$. Substituting $m \approx \log(1/\varepsilon)$, we obtain the dynamic learnability complexity $T(\varepsilon) \approx \frac{m}{\varepsilon} \approx \tilde{O}(1/\varepsilon)$.
\end{proof}

This result establishes that spectral filtering achieves proper \emph{Dynamic Learnability}: it does not merely achieve low average regret, but rather converges to a state of low predictive risk for all future times $t > T(\varepsilon)$.
Crucially, this learnability is achieved \textbf{improperly}: we do not identify the system matrices $A, C$ or the latent state $x_t$. Instead, we learn a predictor in the spectral feature space that is provably sufficient for prediction.
The learnability complexity $T(\varepsilon)$ depends on the number of filters $m$ (which captures the complexity of the Hilbert operator) but remains independent of the state dimension $d$ of the underlying system.

\subsection{Spectral Complexity and Predictability}
\label{sec:spectral_complexity}

The spectral filtering example illustrates a broader principle: whether prediction is possible after
finite observation depends not on identifying the system parameters, nor on the dimension of its latent state $d$,
but on the \emph{effective spectral complexity} of the dynamics.

For symmetric linear dynamical systems, the Bayes–optimal predictor is an infinite convolution whose
coefficients are mixtures of exponentials $\lambda^k$. The ability to approximate this predictor using
a finite number of filters is governed by the decay of the spectrum of the associated Hankel (or Hilbert)
operator. Fast spectral decay implies that only a small number of modes are needed for accurate
prediction, even if the underlying state dimension $d$ is very large.

This perspective extends beyond linear systems. Recent work \cite{Dogariu2025} shows that a wide
class of nonlinear dynamical systems admit finite-dimensional linear observers whose prediction error
decays over time. Concretely, these observers take the form of linear \emph{Luenberger systems} that maintain an
internal state updated from past observations and produce one-step predictions.
Rather than reconstructing the full latent state of the system, the observer is optimized directly
for predictive accuracy.
Among all such linear observers, one can ask for the minimal state dimension required to achieve
prediction error at most $\varepsilon$ uniformly over time.
This minimal dimension defines the spectral complexity parameter $Q^\star(\varepsilon)$.

Crucially, for many systems of interest—such as dissipative physical systems—this predictive dimension
$Q^\star$ is significantly smaller than the ambient state dimension ($Q^\star \ll d$).
Intuitively, $Q^\star$ measures how many independent dynamical modes must be tracked in order to predict
future observations. Systems with small $Q^\star$ rapidly ``forget'' their initial conditions under
observation, enabling finite-time prediction without system identification.

In contrast, systems with large or infinite $Q^\star$—such as chaotic systems where initial information is amplifed by positive Lyapunov exponents—may require arbitrarily long observation horizons to resolve the state, making prediction fundamentally difficult.

This notion explains why spectral filtering succeeds in Section~2.1: it implicitly constructs a
predictor whose dimension scales with the spectral complexity $Q^\star$ rather than with the state dimension $d$ of
the system.
\section{Learnability in Dynamical Systems and Stochastic Processes}
\label{sec:learnability-conditions}

In this section we formalize the notion of {dynamic learnability}.
We begin by defining dynamical systems as a generative model for sequential data,
then show how such systems induce stochastic processes,
and finally state general learnability definitions that apply to
all stationary stochastic processes.

\subsection{Dynamical Systems}
\label{subsec:dynamical-systems}

A discrete--time stochastic {dynamical system} (or partially observed state-space model)
is defined by latent states $(x_t)$ and observations $(y_t)$ evolving as
\begin{align}
x_{t+1} &= f(x_t, w_t), \label{eq:dynamics}\\
y_t &= h(x_t, v_t), \label{eq:observation}
\end{align}
where $(w_t)$ and $(v_t)$ are independent, identically distributed mean-zero noise processes.
Here $f$ is the \emph{transition map} and $h$ the \emph{observation map}.
We will often consider dynamical systems satisfying the following regularity conditions, which are standard in control and filtering and ensure well-posed prediction problems (this is the case for the results in \cite{Dogariu2025}). 

\begin{enumerate}
    \item[A1] \textbf{Bounded trajectories.}
    For all initial states $x_0$, the trajectory $(x_t)$ remains bounded almost surely.
    \item[A2] \textbf{Lipschitz regularity.}
    The maps $f$ and $h$ are $1$-Lipschitz in their first argument, uniformly over noise realizations.
\end{enumerate}

These assumptions, standard in stochastic control, ensure that trajectories are stable and perturbations
do not amplify unboundedly.
When $(w_t)$ and $(v_t)$ are stationary and the system is stable,
the induced output process $(y_t)$ is also stationary. These assumptions are \emph{not required} for the definitions that follow, but will be useful
for discussing positive learnability results. 

A dynamical system together with an initial distribution $P_I$ on $x_0$
and noise laws $P_W, P_V$ defines a probability distribution $P_{f,h,P_I}$
over the infinite observation sequence $(Y_t)_{t\ge0}$.
This distribution is called the {observation process} of the system.

\subsection{Stochastic Processes}
\label{subsec:stochastic-processes}

A (discrete--time) \emph{stochastic process} is a probability distribution
$P$ over infinite observation sequences $(Y_t)_{t\ge0}$,
where each $Y_t$ takes values in a measurable space $\mathcal Y$.
We say that $P$ is \emph{stationary} if it is invariant to time shifts:
for all $h\ge0$ and $k\ge1$,
\[
(Y_{t+1},\ldots,Y_{t+k}) \sim (Y_{t+1+h},\ldots,Y_{t+k+h})
\quad \text{under } P.
\]
Stationarity ensures that statistical properties do not depend on absolute time,
and it allows us to reason about long-run predictive performance.
A process may or may not be \emph{ergodic}; our learnability definitions
require only stationarity.

\paragraph{Dynamical systems as stochastic processes.}
Every dynamical system described in Section~\ref{subsec:dynamical-systems},
together with an initial distribution $P_I$ and noise laws $P_W,P_V$,
induces a stochastic process $P_{f,h,P_I}$ on the observation sequence $(Y_t)$.
If the initial distribution $P_I$ is stationary for the transition kernel of $f$,
the induced process is stationary.
Conversely, any stationary stochastic process can be represented as a
(possibly infinite--dimensional) dynamical system by taking the entire history
$x_t = (Y_1,\ldots,Y_{t-1})$ as its latent state.
Thus, dynamical systems are a \emph{structured subclass} of stochastic processes,
and results stated for processes automatically apply to them.

\paragraph{Examples of model classes.}
The stochastic--process viewpoint allows us to treat a wide range of generative
models under a unified learning framework:
\begin{itemize}
    \item \textbf{Parametric dynamical systems.}
    For a parameterized family of state--space models
    $x_{t+1}=f_\theta(x_t,w_t),\; y_t=h_\theta(x_t,v_t)$
    with parameters $\theta\in\Theta$ and noise laws $P_W,P_V$,
    the model class is
    $\mathcal P = \{P_{\theta,x_0} : \theta\in\Theta,\, x_0\in\mathcal X\}$,
    where $P_{\theta,x_0}$ is the induced observation process.
    If $P_I$ is stationary for each $\theta$, we denote
    $\mathcal P = \{P_{\theta,\mathrm{stat}} : \theta\in\Theta\}$.
    Linear, nonlinear, and controlled systems all fit in this category.
    \item \textbf{Markov and $k$--order processes.}
    The class of $k$--Markov processes, in which
    $P(Y_{t+1}\mid Y_{1:t}) = P(Y_{t+1}\mid Y_{t-k+1:t})$,
    can be represented as finite--memory dynamical systems
    with state $x_t = (Y_{t-k+1},\ldots,Y_t)$.
    Hidden Markov models (HMMs) and autoregressive (AR) processes
    are standard examples.
    \item \textbf{Neural sequence models.}
    Given an architecture such as a recurrent network or Transformer
    parameterized by $\theta\in\Theta$,
    one can define $\mathcal P = \{P_\theta : \theta\in\Theta\}$,
    where $P_\theta$ is the process whose conditionals
    $P_\theta(Y_{t+1}\mid Y_{1:t})$
    correspond to the model’s next-token predictions.
    From this viewpoint, training large language models
    is an instance of dynamic learning over such a class. It is important to note that the stochastic process viewpoint is necessary to represent Transformers, that are not naturally modeled as dynamical systems. %\eh{Nati, see this particular note on transformers...}
    
    \item \textbf{Nonparametric families.}
    More generally, $\mathcal P$ may denote any family of stationary processes
    satisfying structural constraints such as finite predictive dimension,
    bounded mutual information rate, or specific mixing properties. %\eh{examples?} 
\end{itemize}

These examples demonstrate that the process-based formulation
strictly generalizes the dynamical-system perspective:
it encompasses finite--dimensional state--space models,
finite--memory stochastic processes, and expressive
neural architectures for sequence prediction,
all under a single unified notion of learnability.

\subsection{Prediction, Loss, and the Bayes--Optimal Predictor}

Let $\widehat{\mathcal Y}$ denote the prediction space,
which may coincide with $\mathcal Y$
or be a space of distributions over $\mathcal Y$.
Let $\mathcal Y^t$ denote the space of observation sequences of length $t$,
and write $\mathcal Y^* = \bigcup_{t\ge0}\mathcal Y^t$ for the set of all finite
observation histories.
A \emph{predictor} is a measurable function
$\pi:\mathcal Y^* \to \widehat{\mathcal Y}$ producing one-step predictions
$\hat Y_{t+1} = \pi(Y_{1:t})$.
Given a loss function $\ell:\widehat{\mathcal Y}\times\mathcal Y\to\mathbb R_+$,
the conditional expected loss under process $P$ is
$\mathbb E_P[\ell(\hat Y_{t+1}, Y_{t+1}) \mid Y_{1:t}]$.

The \emph{Bayes--optimal predictor} for process $P$ is the rule
\begin{equation}
\pi^\star_P(y_{1:t})
~\in~
\arg\min_{\hat y\in\widehat{\mathcal Y}}
\mathbb E_P[\ell(\hat y,Y_{t+1})\mid Y_{1:t}=y_{1:t}],
\label{eq:bayes-optimal}
\end{equation}
which minimizes the expected one-step loss given the past.
The corresponding minimal conditional loss
is the \emph{Bayes risk}.
For standard losses:
\begin{itemize}
\item Squared loss $\ell(\hat y,y)=\|\hat y-y\|_2^2$:
  $\pi^\star_P(y_{1:t}) = \mathbb E_P[Y_{t+1}\mid Y_{1:t}]$.
\item Absolute loss $\ell(\hat y,y)=|\hat y-y|$:
  $\pi^\star_P(y_{1:t})$ is any posterior-predictive median.
\item Log-loss $\ell(\hat y,y)=-\log \hat y(y)$ for $\hat y\in\Delta(\mathcal Y)$:
  $\pi^\star_P(y_{1:t})=P(Y_{t+1}\mid Y_{1:t})$.

\end{itemize}

\noindent\textbf{Deterministic dynamical systems.}
In the special case of deterministic dynamical systems (where $w_t=0$ and $v_t=0$ in equations \eqref{eq:dynamics} and \eqref{eq:observation}) with a \emph{fixed} initial state $x_0$, the entire observation sequence is predetermined.
The conditional distribution $P(Y_{t+1}\mid Y_{1:t}=y_{1:t})$ is therefore a Dirac delta at the true next value $y_{t+1}$.
Consequently, the Bayes--optimal predictor achieves zero loss,
$\pi^\star_P(y_{1:t}) = y_{t+1}$ and
$\ell(\pi^\star_P(y_{1:t}),y_{t+1}) = 0$,
so the Bayes risk vanishes.

The Bayes predictor represents an ideal observer with full knowledge of $P$.
Dynamic learnability asks whether a single algorithm can approach
its predictive performance after observing a finite portion of the sequence.

\subsection{Definitions of Dynamic Learnability}
\label{subsec:learnability-definition}

We now give precise definitions for dynamic learnability. We begin with the structured case of dynamical systems, which highlights the specific role of the initial state and hidden dynamics. We then generalize this to the broader setting of stationary stochastic processes (covering both realizable and agnostic cases).

\begin{definition}[Dynamic Learnability of Dynamical Systems]
\label{def:ds-learnability}
Let $\mathcal C$ be a class of dynamical systems defined by tuples $(f, h, P_W, P_V)$.
A learning algorithm $\mathcal A$ is said to \emph{learn} $\mathcal C$ with dynamic learnability complexity $T(\varepsilon)$ if, for every system in $\mathcal C$ and \textbf{every admissible initial state} $x_0$, the algorithm satisfies
\begin{equation}
\mathbb E \left[
  \ell(\hat Y_{t+1}, Y_{t+1}) - \ell(\pi^\star, Y_{t+1})
\right] \le \varepsilon
\label{eq:ds-learnability}
\end{equation}
for all $t \ge T(\varepsilon)$.
\end{definition}

This definition explicitly captures the requirement that the learner must resolve the uncertainty of the hidden state $x_t$ (induced by $x_0$ and the noise history) within finite time $T(\varepsilon)$.

To capture broader model classes, such as large language models, we extend this to stationary processes.

\begin{definition}[Realizable (well-specified) dynamic learnability]
\label{def:realizable-learnability}
Let $\mathcal P$ be a family of stationary stochastic processes on $\mathcal Y$,
and let $\ell:\widehat{\mathcal Y}\times\mathcal Y\to\mathbb R_+$ be a loss function.
A prediction algorithm $\mathcal A$ producing $\hat Y_{t+1}=\mathcal A(Y_{1:t})$
is said to \emph{learn} $\mathcal P$ with dynamic learnability complexity $T(\varepsilon)$
if for every $\varepsilon>0$ there exists $T(\varepsilon)<\infty$ such that
for all $t>T(\varepsilon)$ and every $P\in\mathcal P$,
\begin{equation}
\mathbb E_{P}\!\left[
  \ell(\hat Y_{t+1},Y_{t+1})
  - \ell(\pi^\star_P(Y_{1:t}),Y_{t+1})
\right]
~\le~ \varepsilon.
\label{eq:process-realizable}
\end{equation}
\end{definition}

The burn-in $T(\varepsilon)$ quantifies how long the learner must observe
before achieving expected loss within $\varepsilon$ of the Bayes predictor.

\begin{definition}[Agnostic dynamic learnability]
\label{def:agnostic-learnability}
Let $\Pi$ be a class of predictors $\pi:\mathcal Y^*\!\to\!\widehat{\mathcal Y}$.
An algorithm $\mathcal A$ is said to \emph{learn} $\Pi$ with dynamic learnability complexity 
$T(\varepsilon)$ if for every $\varepsilon>0$, every stationary process~$P$, and all $t>T(\varepsilon)$,
\begin{equation}
\mathbb E_{P}\!\left[
  \ell(\hat Y^{\mathcal A}_{t+1},Y_{t+1})
  - \inf_{\pi\in\Pi}
    \ell(\pi(Y_{1:t}),Y_{t+1})
\right]
~\le~ \varepsilon.
\label{eq:process-agnostic}
\end{equation}
\end{definition}

The realizable case corresponds to $\mathcal P=\{P_{f,h,P_I}\!:\!(f,h,P_I)\in\mathcal F\}$,
the class of processes induced by dynamical systems in~$\mathcal F$,
while the agnostic case compares the learner to the best predictor in a given class~$\Pi$.

These definitions generalize classical PAC and online learning notions.
If $\mathcal P$ consists of i.i.d.\ processes, they reduce to PAC learnability;
if $\mathcal P$ includes all deterministic sequences, they coincide with
online learnability and small regret.
For dynamical systems, they capture the ability to resolve hidden-state
uncertainty from finite observations.
The \emph{Dynamic Learnability Complexity} $T(\varepsilon)$ thus measures
how long a learner must observe the process before accurate one-step
prediction becomes possible, serving as a sequential analogue
of statistical sample complexity.

\subsection{Examples}
\label{sec:examples}

To ground the discussion, we illustrate the notion of dynamical learnability with representative examples spanning modern AI, physics, and control. 
In each case, the example implicitly defines a family of stochastic processes $\mathcal P$, indexed by an initial state or initialization, and the dynamic learnability complexity $T(\varepsilon)$ captures the time required to resolve latent-state uncertainty well enough for reliable next-step prediction.

\paragraph{Controlled dynamical system with a fixed policy.}
Consider a linear system under a stabilizing linear feedback policy $K$, see, e.g., \cite{Ljung1998,hazan2022introduction}:
\begin{align*}
    x_{t+1} &= Ax_{t}+Bu_{t} \\
    u_{t} &= Kx_{t}.
\end{align*}
The closed-loop dynamics reduce to $x_{t+1}=(A+BK)x_{t}$. If $(A+BK)$ is stable, then trajectories remain bounded and the influence of the initial state decays over time.

In this case, the family $\mathcal P$ consists of stable linear stochastic processes indexed by the initial condition $x_0$. 
Stability ensures contraction of hidden-state uncertainty, implying the existence of a finite dynamic learnability complexity $T(\varepsilon)$ beyond which accurate next-step prediction is possible uniformly in time.
Importantly, learnability does not require identification of the system matrices, but only prediction of future observations.

More generally, in reinforcement learning environments, once a policy is fixed, the agent--environment interaction defines a dynamical system which may be dynamically learnable in this sense.

\paragraph{The Lorenz attractor.}
The Lorenz system \cite{Lorenz1963},
\begin{align*}
    \dot{x} &= \sigma(y-x) \\
    \dot{y} &= x(\rho-z)-y \\
    \dot{z} &= xy-\beta z
\end{align*}
is a prototypical chaotic dynamical system. Even small errors in estimating the state grow exponentially with time, making long-horizon prediction essentially impossible.

From the perspective of our framework, each initial condition $x_0$ induces a distinct (deterministic) stochastic process $P_{x_0} \in \mathcal P$. 
While the Bayes-optimal predictor $\pi^\star_P$ achieves zero one-step loss, the exponential sensitivity to initial conditions implies that resolving the hidden state may require an arbitrarily long observation horizon. 
This illustrates a fundamental obstruction to dynamic learnability that is intrinsic to the system, rather than to any particular learning algorithm.

Some recent methods have nevertheless shown promise for next-token prediction in chaotic systems without identifying the underlying dynamics, focusing instead on short-horizon dynamic learnability in the sense studied here \cite{Dogariu2025}.

\paragraph{Fluid Dynamics and Weather Forecasting.}
A prime example is forecasting fluid flow, which underpins weather and climate modeling. This involves learning the evolution of systems governed by partial differential equations (PDEs) such as the incompressible Navier-Stokes equations:
\begin{align*}
\frac{\partial \mathbf{u}}{\partial t} + (\mathbf{u} \cdot \nabla) \mathbf{u} &= -\frac{1}{\rho} \nabla p + \nu \nabla^2 \mathbf{u} \\
\nabla \cdot \mathbf{u} &= 0
\end{align*}
where $\mathbf{u}$ is the velocity field, $p$ is pressure, $\rho$ is density, and $\nu$ is kinematic viscosity.
We cast this physical system into our framework ($x_t, y_t, f$) via spatial discretization (e.g., on a grid or spectral basis) and temporal integration:
\begin{itemize}
    \item The latent state $x_t \in \mathbb{R}^d$ represents the discretized velocity and pressure fields at time step $t$.
    \item The transition map $f$ is the operator that integrates the PDE forward in time by $\Delta \tau$. That is, $x_{t+1} \approx x_t + \int_{t}^{t+\Delta \tau} \text{NS}(x(s)) ds$. The process noise $w_t$ accounts for numerical discretization errors or unresolved sub-grid turbulence.
    \item The observations $y_t = h(x_t)$ represent sparse sensor readings (e.g., weather stations, satellites) rather than the full state.
\end{itemize}
The prediction task is to predict the next sensor reading $y_{t+1}$ given the history of sparse observations $y_{1:t}$, without necessarily having full access to the initial condition $x_0$.

In this context, $\pi^\star_P$ corresponds to a perfect physical simulator (Direct Numerical Simulation) initialized with the optimal posterior estimate of the state given the noisy sensor history. In chaotic regimes (like the atmosphere), comparing to the Bayes predictor allows us to distinguish between model error (deficiencies in the learner) and fundamental chaos (Lyapunov instability limiting the horizon).

This framework captures the constraints of physics better than the alternatives.
Unlike PAC learning, which treats weather maps as independent snapshots (asking ``can I recognize a storm?''), dynamic learnability addresses the \emph{data assimilation} problem: resolving the unobserved velocity field $x_t$ from the sequence $y_{1:t}$ well enough to simulate future motion.
Unlike online learning, which assumes the observation sequence can be adversarial, dynamic learnability exploits the fact that physical systems are constrained by conservation laws (mass, momentum) and regularity. Specifically, the viscous term $\nu \nabla^2 \mathbf{u}$ ensures the transition map $f$ is Lipschitz (smoothing), preventing arbitrary adversarial divergence. Because $f$ follows physical laws rather than an adversary, dynamic learnability can potentially provide tighter, physically meaningful guarantees.

\paragraph{Hypothesis: Large Language Models (LLMs) and Training Dynamics.}
A large language model can be viewed as a dynamical system where the hidden state $s_t=(x_t,\theta_t)$ comprises two distinct components: the activation state $x_t$ (e.g., KV cache) representing the current context, and the parameter state $\theta_t$ representing the network weights. The transition function updates both: $x_t$ evolves autoregressively via the architecture (e.g., Transformer block), while $\theta_t$ evolves via the optimization algorithm (e.g., an adaptive gradient method). A precise formulation of how to fit LLMs into dynamic learnability is left for future work.

\section{A Research Program: Future Directions}
\label{sec:future}

The purpose of this note is to propose a research program of learnability in dynamical systems based on the notion of next token prediction. Numerous open directions and investigation areas follow:

\begin{itemize}

    \item \textbf{Characterizing Dynamical Learning Complexity.} While we outlined necessary and sufficient conditions for dynamical learnability, this barely scratches the surface.  A key challenge is to characterize the complexity of dynamic learnability in quantitative terms, such as the Optimal Luenberger Program parameter $Q^\star$. from \cite{Dogariu2025}. What is the relationship between $Q^\star$ and established system properties, such as spectral radii or Lyapunov exponents? How does the spectral complexity from \cite{Dogariu2025} relate to optimal learnability? 

    How do the hidden dimension of the underlying dynamical system, or observable dimension, enter the possible rates? Are there natural transitions in terms of learnability rates such as for statistical learnability \cite{bousquet2021theory}?

   \item \textbf{Algorithmic efficiency.} Designing efficient (polynomial-time) learners remains, in our opinion, the most important open challenge.

Indeed, designing provably efficient methods for nonlinear dynamical systems is an important open problem in all of control and reinforcement learning, with only few positive examples \footnote{such as Tabular MDPs and Linear MDPs}. A notable exception is that of linear dynamical systems,  for which the canonical work of Kalman established the LQR, see references in e.g. \cite{hazan2022introduction}.   

    One important advancement is the method of spectral filtering, which was recently shown to give provable dynamical learnability even for nonlinear dynamical systems \cite{Dogariu2025}. 

    \item \textbf{Equilibria arising from dynamic learnability.} The ability to predict and learn in the dynamical systems gives rise to natural questions in game theory. What solution concepts arise from dynamical learning in multiplayer games? Are these efficiently attainable by independent players? How are spectral properties of the game affected by natural strategies, and what convergence rates do they allow?

    \item \textbf{Average-case vs. worst-case learnability in terms of the initial state.} Relax the uniform-in-time, worst-case-over-$x_0$ definition of learnability. This involves asking whether \emph{typical} trajectories are identifiable under probabilistic assumptions and connecting this to classical statistical complexity measures. This is  particularly important for chaotic systems such as the Lorenz system.

    \item \textbf{Time-varying dynamical systems.} Our model assumes a stationary dynamical system, but we know that the real world is rarely stationary. The datasets used to train LLMs were created by different authors, or ``brains", each with its own separate dynamics. Similarly, the observation function may change over time due to change of perspective, or machinery, or change of language. 
    
    How robust is dynamical learnability to change in dynamics or observation? Are abrupt changes harder to handle than gradual? The machinery from adaptive and dynamic regret minimization seems useful for this investigation. 

    \item \textbf{Learnability in Active Control and Reinforcement Learning.} The paper intentionally excludes open-loop inputs and considers only fixed-policy control as a positive example. We believe there is plenty to investigate in this variant, and there are reasons to believe that open-loop inputs render learnability much more challenging.  The reason is that open loop according to a general dynamical controller can completely change a given dynamical system to any other dynamical system.  
    
    That said, a major open direction is to study dynamic learnability within an active control loop, such as in reinforcement learning. How does the sample complexity $T(\epsilon)$ behave when the agent's policy is non-stationary, thereby actively changing the system dynamics over time  \cite{kakade2003sample,foster2022complexity}?

    \item \textbf{Why does next-token prediction give rise to intelligent behavior?}  This is the question we started with, and studying dynamic learnability may give us hints in the direction of a solution. Under which circumstances does dynamic learnability imply reconstruction of the underlying dynamics?  Can this hold even with improper learning such as spectral filtering methods?

\end{itemize}

\bibliographystyle{plain}
\bibliography{main}

\end{document}